\documentclass[lettersize,journal]{IEEEtran}
\usepackage{amsmath,amsfonts}
\usepackage[ruled,linesnumbered]{algorithm2e}
\usepackage{array}
\usepackage{textcomp}
\usepackage{stfloats}
\usepackage{url}
\usepackage{verbatim}
\usepackage{graphicx}
\usepackage{cite}
\usepackage{booktabs}
\usepackage{threeparttable}
\usepackage{pifont}
\usepackage{ragged2e} 
\usepackage {graphicx}
\usepackage{booktabs,makecell, multirow, tabularx}
\usepackage{subfigure}

\usepackage{enumitem}
\usepackage{bm}
\usepackage{amssymb}
\usepackage{xcolor}
\usepackage{soul}
\sethlcolor{yellow}

\soulregister\cite7
\soulregister\ref7 
\soulregister\eqref7 
\soulregister\pageref7 
\begin{document}

\title{RHAML: Rendezvous-based Hierarchical Architecture for Mutual Localization}

\author{Gaoming Chen, Kun Song, Xiang Xu, Wenhang Liu, Zhenhua Xiong, \IEEEmembership{Member, IEEE}
\thanks{
	This work was in part supported by the National Natural Science Foundation of China (U1813224) and MoE Key Lab of Artificial Intelligence, AI Institute, Shanghai Jiao Tong University, China. \textit{(Corresponding author: Zhenhua Xiong.)}
}
\thanks{
	Gaoming Chen, Kun Song, Xiang Xu, Wenhang Liu, and Zhenhua Xiong are with the School of Mechanical Engineering, Shanghai Jiao Tong University, Shanghai, China (e-mail: cgm1015@sjtu.edu.cn; coldtea@sjtu.edu.cn; xu-xiang@sjtu.edu.cn; liuwenhang@sjtu.edu.cn; mexiong@sjtu.edu.cn).
}
}
\maketitle

\begin{abstract}
Mutual localization serves as the foundation for collaborative perception and task assignment in multi-robot systems.
Effectively utilizing limited onboard sensors for mutual localization between marker-less robots is a worthwhile goal.
However, due to inadequate consideration of large scale variations of the observed robot and localization refinement, previous work has shown limited accuracy when robots are equipped only with RGB cameras.
To enhance the precision of localization, this paper proposes a novel rendezvous-based hierarchical architecture for mutual localization (RHAML).
Firstly, to learn multi-scale robot features, anisotropic convolutions are introduced into the network, yielding initial localization results.
Then, the iterative refinement module with rendering is employed to adjust the observed robot poses.
Finally, the pose graph is conducted to globally optimize all localization results, which takes into account multi-frame observations.
Therefore, a flexible architecture is provided that allows for the selection of appropriate modules based on requirements.
Simulations demonstrate that RHAML effectively addresses the problem of multi-robot mutual localization, achieving translation errors below 2 cm and rotation errors below 0.5 degrees when robots exhibit 5 m of depth variation.
Moreover, its practical utility is validated by applying it to map fusion when multi-robots explore unknown environments.
\end{abstract}


\section{Introduction}
\IEEEPARstart{M}{utual} localization between multiple robots can convert the perception of each robot's local reference frame to a common reference frame, which is the foundation of many advanced tasks, such as collaborative exploration\cite{CMU_explor}, surveillance\cite{Pasqualetti}, and transportation\cite{Hu}.
In environments where GPS-denied or motion capture systems are unavailable, such as large indoor scenes, the above tasks become more difficult.
One solution is for robots to have known initial poses\cite{Dong}.
However, this method relies too heavily on priori hypotheses and is prone to failure in complex scenes\cite{Gaofei2022}.

Only using limited onboard sensors for mutual localization between marker-less robots is more attractive.
According to the principles of different sensors, mutual localization methods primarily include two categories, i.e., distance-based localization and bearing-based localization\cite{Zhou}.
The first category of methods requires the robot to be equipped with specific sensors such as UWB\cite{Liu}, which are dedicated solely to the single task of localization, increasing the cost of the system.
This method will not be considered in subsequent content.
In contrast, bearing-based methods can rely on sensors, such as RGB cameras\cite{Arcuo_1}, commonly present in advanced tasks, which can obtain relative pose without increasing the number of sensors.
Consequently, it is of great significance to achieve mutual localization with affordable cameras.

Typically, only relying on visual information, one solution is to incorporate interloop detection\cite{CCM-SLAM, TopoMap, Kimera-Multi} into multi-robot Simultaneous Localization and Mapping (SLAM).
It means that robots have observed the same scene.
However, for scenes with weak textures or high repeatability, this method is prone to failure.
Another commonly adopted method is based on robot rendezvous\cite{CNN_trans}, which means that the observed robot appears in the observer robot's field of view (FOV).
This method only focuses on the robot itself and is basically not affected by the external environment.
When marker-less robots rendezvous, previous methods have utilized deep learning\cite{DL_Rendezvous2,DOPE,Bultmann} for mutual localization.
However, the use of general convolutional kernels makes it challenging to address the large scale variations caused by the robot motion, and inadequate consideration of refinement further limits the accuracy.
\begin{figure*}[!t]
	\centering
	\includegraphics[width=6.5in]{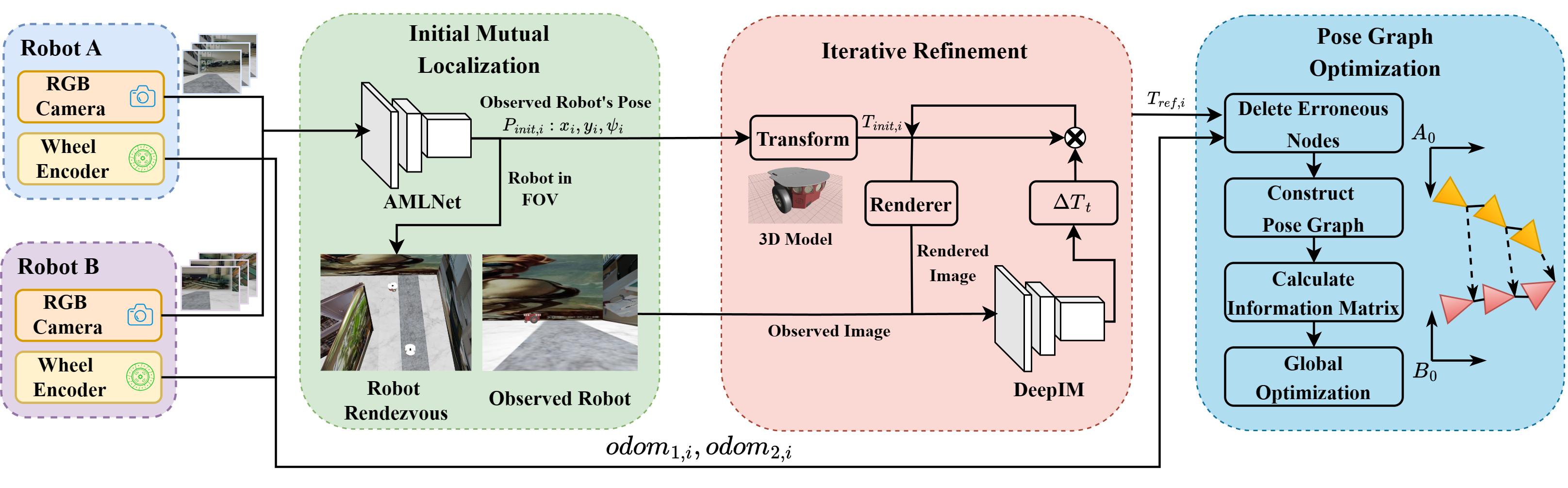}
	\vspace{-2mm}
	\caption{Overview of the proposed architecture for multi-robot mutual localization. Team robots send their respective RGB images and wheel odometry to the central node. Firstly, the Initial Mutual Localization module inputs the images captured by robots into our AMLNet to detect whether an observed robot is in the FOV, and outputs the initial localization result ${P}_{init,i}$ to the Iterative Refinement module. Then, by iteratively rendering the 3D model, the refined localization ${T}_{ref,i}$ is obtained through DeepIM. After that, the selected localization results are optimized by constructing the pose graph.}
	\label{architecture}
	\vspace{-4mm}	
\end{figure*}

Thus, a novel \textbf{R}endezvous-based \textbf{H}ierarchical \textbf{A}rchitecture for \textbf{M}utual \textbf{L}ocalization (RHAML) is proposed in this paper.
Our system takes RGB images and odometry as inputs, aiming to achieve high-precision mutual localization through a hierarchical framework that is decoupled from the external environment.
Firstly, we propose an Anisotropic Mutual Localization Network (AMLNet) that predicts the presence of the robot and its pose in the FOV.
The backbone is augmented with anisotropic convolutional modules, which allow for flexible receptive fields and effectively handle large scale variations (Section~\ref{IML}).
Then, by comparing the rendered image with the observed image, the initial mutual localization result can be iteratively refined with DeepIM\cite{li2018deepim}, leading to higher localization accuracy (Section~\ref{IR}).
Finally, by combining odometry and refined results over a period of time, the pose graph optimization (PGO) is employed to globally adjust all poses.
This is achieved by removing outliers and minimizing the localization errors, resulting in more accurate robot orientation (Section~\ref{PGO}).
The main contributions are as follows:
\begin{itemize}[leftmargin=*]
	\item RHAML is proposed for marker-less multi-robot mutual localization. The hierarchical architecture achieves high accuracy by sequentially fine-tuning the pose and allows flexible configuration selection according to specific needs.
	\item AMLNet is proposed for initial mutual localization, which does not require any hand-crafted image features and handles large scale variations through anisotropic convolutions.
	\item Extensive simulations and experiments demonstrate that RHAML achieves high-precision localization with translation errors below 2 cm and rotation errors below 0.5 degrees.
\end{itemize}
\section{Related Work}
Recently, bearing-based mutual localization methods have gained widespread attention.
Depending on different approaches, they can be primarily divided into two categories: interloop detection-based and rendezvous-based.
\subsection{Interloop Detection-based Mutual Localization}
Most research related to multi-robot SLAM focuses on interloop detection-based mutual localization, which can be achieved by performing multi-frame matching between robots.

In \cite{CCM-SLAM,TopoMap}, features are extracted for each keyframe.
When similar features are found, loop closures are constructed.
In \cite{Kimera-Multi}, mutual localization is achieved through geometric verification, and a robust PGO solver is proposed to handle outliers.
The hybrid descriptor that combines both semantic and geometric information is proposed in \cite{Semanticgeometric}, and the system's robustness is improved through a local score enhancement strategy.
A new position descriptor is developed based on spherical projection and multiperspective fusion \cite{AutoMerge}.
Meanwhile, active sequence association is used to reduce the length of overlapping sequences.

Despite the progress made by the above methods, robust mutual localization relies on multi-frame associations, which is time-consuming.
Moreover, in scenes with weak textures or high repetition, even if robots have highly overlapping paths, mutual localization remains a challenging problem.
\subsection{Rendezvous-based Mutual Localization}
To tackle these challenges, recent research has emphasized rendezvous-based methods, where the localization of the observed robot is directly determined through observations.

In \cite{Arcuo_2,Gaofei2022,Gaofei2023}, AprilTag is used for bearing measurement.
In \cite{Arcuo_1}, localization results are obtained by recognizing non-static features in the scene.
In \cite{CNN_xyzBouning1,CNN_xyzBouning2}, the convolutional neural network (CNN) is employed to detect MAV in images, and the position is calculated based on the size of the bounding box combined with the actual dimensions.
In \cite{DL_Rendezvous1}, CNN is utilized to directly predict the position of the robot.
Afterward, although the orientation is integrated in \cite{DL_Rendezvous2}, the inference speed is fast but the learned feature is limited because of its lightweight model.
\cite{UAV_CNN} further optimize the CNN, achieving high-precision positioning of MAV.
In \cite{Bultmann}, CNN is used to detect keypoints directly and PGO is adopted to refine poses.
In \cite{CNN_trans}, DOPE \cite{DOPE} is utilized to infer the 2D image coordinates of projected 3D bounding boxes, but it exhibits some outliers.
In \cite{PVNet}, keypoint localization is achieved through a vector-field representation.
The accuracy for small objects is limited due to the scarcity of effective pixels.
In \cite{OnePose}, pose estimation is achieved without the need for CAD models.
Instead, object representation is obtained through pointcloud reconstruction.
\cite{CosyPose} adopts multi-frame observations to predict object poses, based on the premise that the target object is stationary.
Anisotropic convolution is first introduced in \cite{Anisotropic}, and it has shown excellent performance in semantic scene completion.
DeepIM\cite{li2018deepim} refines the pose for each frame through iterative rendering, achieving high localization accuracy by comparing observed images with rendered images.

Theoretically, the above methods have achieved mutual localization.
However, for marker-less robots, the accuracy of localization through learning-based methods is limited.
Therefore, RHAML is proposed to achieve high-precision mutual localization through hierarchical optimizations.
\section{Methodology}
The framework of RHAML is shown in Fig.~\ref{architecture}, taking RGB images and wheel encoders as inputs.
Firstly, AMLNet is utilized to extract image features and determine the presence of robots, achieving initial mutual localization.
Then, DeepIM \cite{li2018deepim} is applied to iteratively refine the pose through rendering.
Finally, by conditionally selecting nodes, the pose graph is constructed to globally optimize the results.
\subsection{Initial Mutual Localization}
\label{IML}
Leveraging AMLNet for initial mutual localization (IML), it takes images captured by robots as input and extracts features through multi-layer CNN.
AMLNet finally regresses ${\hat o}_{i}$, ${\hat x}_{i}$, ${\hat y}_{i}$, ${\hat \psi}_{i}$, where $\hat{o}_{i}$ represents whether the robot is within the FOV of the observer robot, indicating whether robots rendezvous, and ${P}_{init,i}=\{\hat{x}_{i}, \hat{y}_{i}, \hat{\psi}_{i}\}$ denotes its pose.

\begin{figure}[t]
	\centering 
	\subfigure[]{
		\centering 
		\includegraphics[height=1.92in]{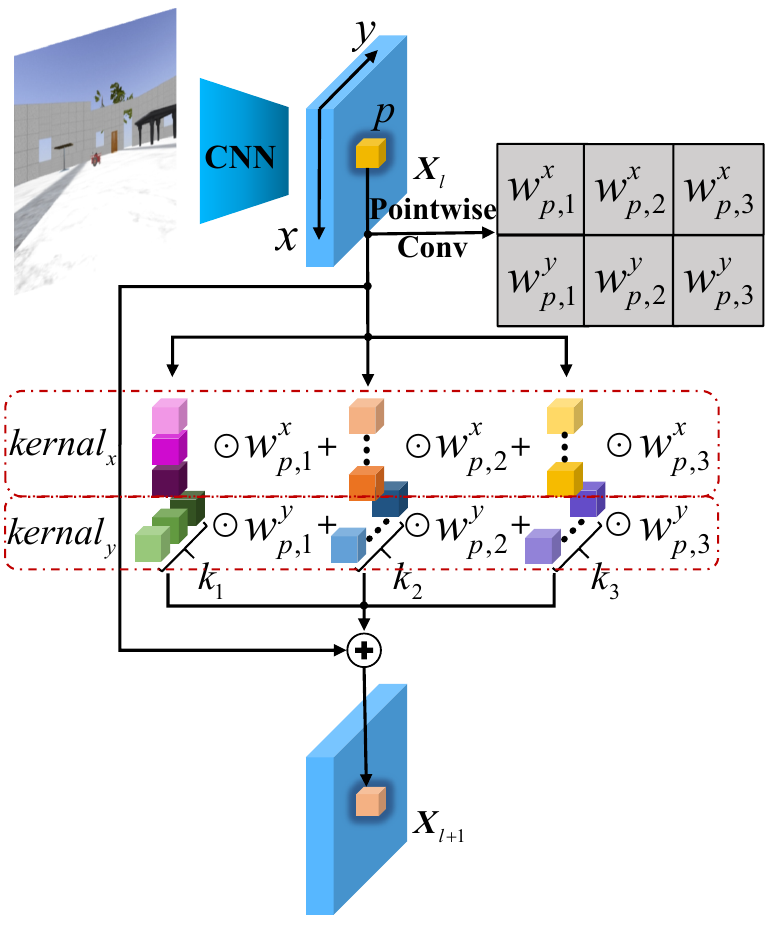}
		\label{ACN}
	}
	\subfigure[]{ 
		\centering   
		\includegraphics[height=1.92in]{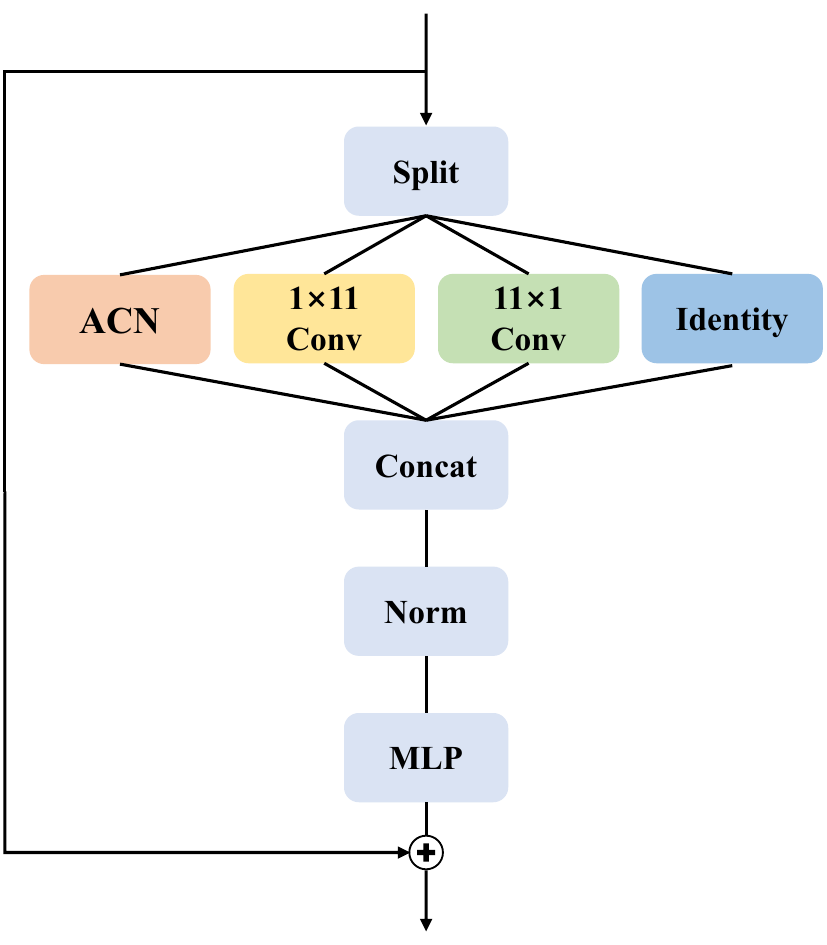}
		\label{AIncepblock}
	}
	\caption{Illustration of ACN and AIncep block. (a) ACN: Each dimension of the feature map is calculated by multiple convolutional kernels of different sizes, and their weights are learned through training. (b) The structure of the AIncep block, including the ACN module.}   
	\label{ANNfig}
	\vspace{-5mm}
\end{figure}

Due to the large motion space of mobile robots, the corresponding pixel regions in the image vary in size.
Towards the goal of adapting to the impact of this, inspired by \cite{Anisotropic}, we introduce Anisotropic Convolution (ACN) in the 2D CNN to achieve adaptive adjustment of convolution kernels when the robot has large scale changes.
Fig.~\ref{ACN} illustrates the structure of ACN.
Unlike the traditional convolutional kernel with the size of ${k}\times{k}$, ACN firstly decomposes the image in the $x$ and $y$ directions and employs ${1}\times{k}$ and ${k}\times{1}$ unidirectional convolutional kernels instead, which significantly reduces the number of parameters.
To achieve the flexible receptive field, ACN then utilizes $n$ unidirectional convolutional kernels of different sizes in ${\mathcal{K}}=\{k_1,k_2,\cdots,k_n\}$ in each dimension of the feature map, which can achieve multi-scale perception and capture varying context information with approximate parameter quantities.
By learning the weights of these convolutional kernels, ACN dynamically adjusts the receptive field size.

Given the input of ACN as $\boldsymbol{{X}}_{l} \in {{\mathbb{R}}^{C\times W\times H}}$, which represents the $l$-th feature map, for the pixel $p$, the weights of the corresponding anisotropic convolution kernels, denoted as $\boldsymbol{\omega}_{p}=\{{\omega_{p,1}^{x}},{\omega_{p,1}^{y}},\cdots, {\omega_{p,n}^{x}},{\omega_{p,n}^{y}}\}$, are learned through point-wise convolution: 
\begin{equation}
	\label{eq1}
	{\boldsymbol{\omega }_{p}}=f({\boldsymbol{X}_{l}}[p],\zeta ) \
\end{equation}
where $f$ represents point-wise convolution operation, and ${\zeta}$ is the parameter of the convolution kernel to be learned.
Therefore, the receptive field can be dynamically adjusted independently based on different weights for each pixel.
By applying Softmax, the weights for each direction are ensured to be greater than 0 and their sum is equal to 1.
Combining residual design, the output of ACN can be denoted as follows:
\begin{equation}
	\label{eq2}
	{\boldsymbol{X}_{l+1}}[p]=\sum\limits_{u\in \{x,y\}}{\sum\limits_{v=0}^{n}{\omega _{p,v}^{u}g({\boldsymbol{X}_{l}}[p], k_v, {\phi} _{v}^{u})+}}{\boldsymbol{X}_{l}}[p] \
\end{equation}
where $g$ represents unidirectional convolution operation, $k_v$ represents the size of the $v$-th unidirectional convolutional kernel, and ${\phi} _{v}^{u}$ is the parameter to be learned.

Benefiting from the excellent feature extraction capability and high training throughputs of the InceptionNeXt block mentioned in \cite{yu2023inceptionnext}, we propose the Anisotropic InceptionNeXt (AIncep) block.
As shown in Fig.~\ref{AIncepblock}, we retain the original identity mapping, $1\times11$ depthwise convolution, and $11\times1$ depthwise convolution due to the good performance of large convolutional kernels.
The difference is that the ACN module replaces the $3\times3$ convolutional kernel, which allows a more diverse receptive field in the entire block.
\begin{figure}[t]
	\centering
	\includegraphics[height=1.3in]{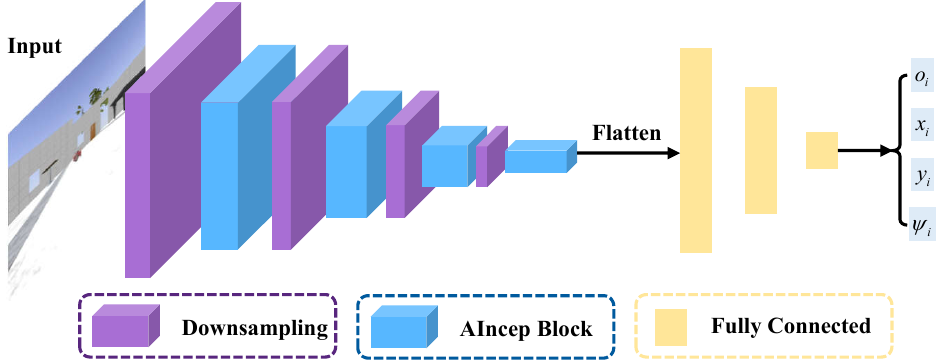}
	\caption{The architecture of AMLNet primarily consists of downsamplings, AIncep blocks, and fully connected layers. It is designed to extract features from the images captured by the observer robot and predict ${o}_{i}$ and ${P}_{init, i}$.}
	\label{AMLNet}
	\vspace{-4mm}
\end{figure}

Finally, we propose AMLNet for predicting the localization of the observed robot.
The backbone of AMLNet is similar to \cite{yu2023inceptionnext}, with the difference of the replacement of InceptionNeXt blocks with AIncep blocks.
Two fully connected (FC) layers are applied as the neck.
In the end, a FC layer serves as the output head to predict $\hat{o}_{i}$ and ${P}_{init, i}$.
Fig.~\ref{AMLNet} shows the framework of AMLNet, which primarily consists of downsamplings, AIncep blocks, and FC layers.
By performing four iterations of downsampling and AIncep blocks, the encoded features are flattened and finally predicted the output through FC layers.

Given the batch size $N$, the goal of AMLNet is to predict $\hat{o}_{i}$ and ${P}_{init, i}$ as closely as possible to the ground truth.
To supervise the presence of the observed robot in the image, we define the observation loss ${L}_{o}$ leveraging Binary Cross Entropy (BCE) loss as follows:
\begin{equation}
	\label{eq3}
	{{L}_{o}}=-\frac{1}{N}\sum\limits_{i=1}^{N}{[{{o}_{i}}\cdot \log ({{{\hat{o}}}_{i}})+(1-{{o}_{i}})\cdot \log (1-{{{\hat{o}}}_{i}})]}\
\end{equation}
where ${o}_{i}=1$ represents the presence of a robot in the FOV of the observer robot, and conversely, ${o}_{i}=0$ denotes its absence.
We define the localization loss ${L}_{p}$ adopting Mean Squared Error (MSE) loss to supervise the pose of the observed robot:
\begin{equation}
	\label{eq4}
	{{L}_{p}}=\frac{1}{N}\sum\limits_{i=1}^{N}{[\sum\limits_{u\in \{x,y\}}{{{({{u}_{i}} - {{{\hat{u}}}_{i}})^{2}}}}}+{{\omega }_{\psi }}{{({{\psi }_{i}} - {{\hat{\psi }}_{i}})^{2}}}]\
\end{equation}
where ${\omega }_{\psi }$ is the weight for the loss of yaw angle ${\psi }_{i}$.
Combining the two parts mentioned above, the overall loss function ${L}_{A}$ of AMLNet is as follows:
\begin{equation}
	\label{eq5}
	{L}_{A}={{L}_{o}}+{{\omega }_{p}}\cdot {{L}_{p}}\
\end{equation}
where ${\omega }_{p}$ is the weight for the localization loss.

Generally, AMLNet can predict whether a robot exists within the FOV of the observer robot and estimate its pose with only an image.
Currently, it can handle scenes where just one robot is present in the FOV.
AMLNet does not need to explicitly establish 2D-3D correspondences for keypoints but directly regresses the pose from the image.
However, due to the large scale variations, the accuracy of directly regressing the pose from images for localization is limited, requiring further refinement in the subsequent process.
\subsection{Iterative Refinement}
\label{IR}
When robots appear in the FOV of the observer robot, local iterative refinement (IR) with ${P}_{init, i}$ can be applied to improve the accuracy of mutual localization during robot rendezvous.
Given ${P}_{init, i}$ and the 3D model of the robot, a relative SE(3) transformation can be calculated by matching the rendered image with the real image captured by the observer robot.
\begin{figure}[t]
	\centering
	\includegraphics[height=1.5in]{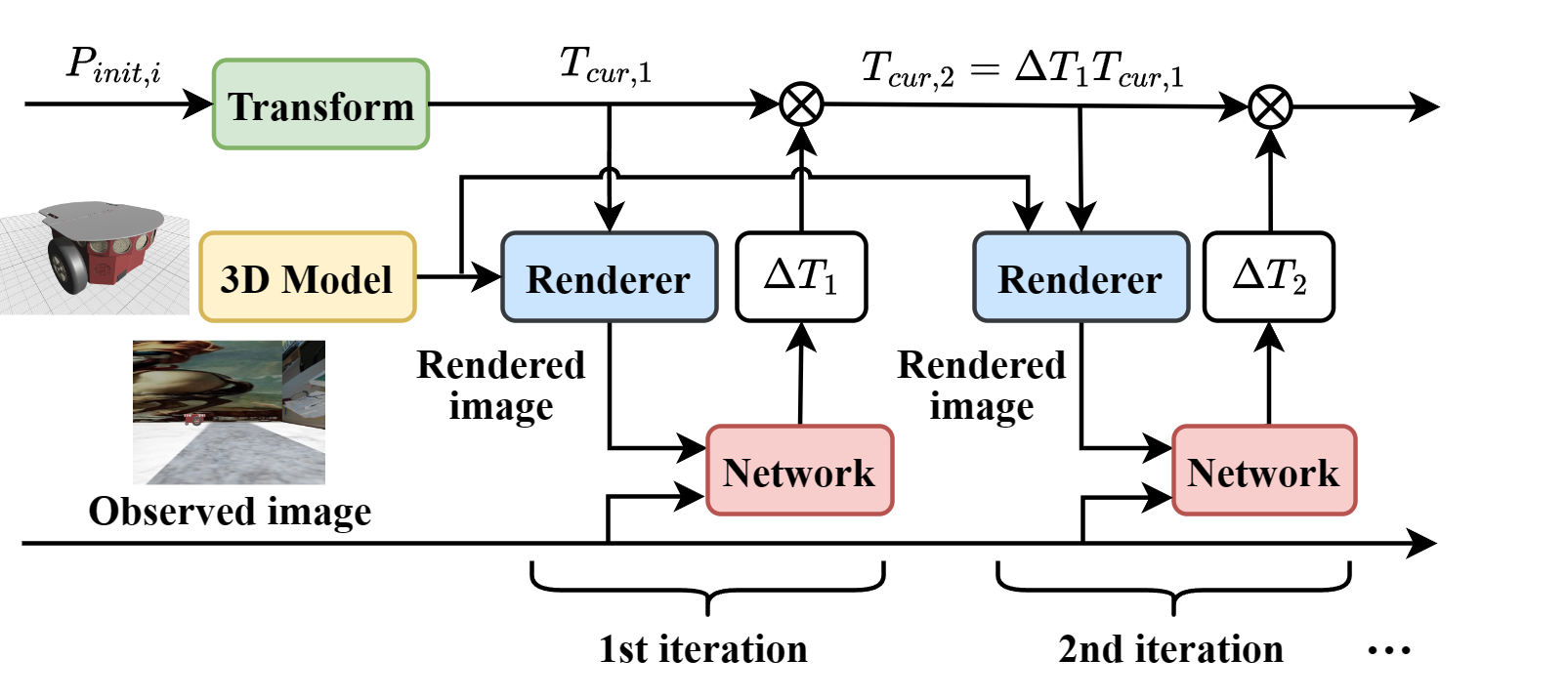}
	\vspace{-2mm}
	\caption{The architecture of the iterative refinement module. Based on the initial mutual localization result and the CAD model of the robot, the image is rendered. Combined with the observed image, they are input into the network to output the relative transformation. Through an iterative loop, refined localization is finally obtained.}
	\label{deepim}
	\vspace{-4mm}
\end{figure}

As shown in Fig.~\ref{deepim}, our IR module is based on DeepIM\cite{li2018deepim} because of its excellent performance in pose refinement.
Firstly, transform ${P}_{init, i}$ to ${T}_{init, i}$ in SE(3).
Using ${T}_{init, i}$ as the initial transformation, the robot's CAD model is rendered to obtain the rendered image.
Then, the rendered image, observed image, and their respective masks are concatenated and fed into the network.
The network leverages CNN layers to encode the image features and outputs the relative transformation ${\Delta}T_{t}$ between the current predicted pose ${T}_{cur, t}$ and the ground truth ${T}_{gt,i}$.
By applying the new transformation ${T}_{cur,t+1}$ to render the CAD model using Eq. \eqref{eq7}, an iterative loop is performed to refine the pose, ultimately resulting in a more accurate ${T}_{ref, i}$.
\begin{equation}
	\label{eq7}
	{{T}_{cur,t+1}}=\left\{ \begin{aligned}
		&{{T}_{init,i}},&t=0  \\
	    &{\Delta {{T}_{t}}\cdot {{T}_{cur,t}}},&t>0  \\
	\end{aligned} \right.
	\
\end{equation}

To ensure accurate point-to-point correspondence in 3D, the point matching loss ${L}_{pm}$ defined in each iteration is as follows:
\begin{equation} 
	\label{eq8}
	{{L}_{pm}}=\frac{1}{M}\sum\limits_{j=1}^{M}{{{\left\| {{T}_{cur,t}}\cdot {{p}_{j}}-{{T}_{gt,i}}\cdot {{p}_{j}} \right\|}_{1}}}\
\end{equation}
In Eq. \eqref{eq8}, $M$ represents the number of sampled points, which are uniformly sampled on the surface mesh, and ${p}_{j}$ denotes the coordinates of each sampled point.
To accelerate rendering, the shader program is simplified to include only vertex and fragment shaders, excluding lighting calculations and shadow effects.
To leverage the complementary advantages of information sharing in multi-task learning, two auxiliary branches are integrated into the network's training process.
These branches calculate the mask loss and optical flow loss, respectively.
The mask loss is calculated using the sigmoid cross-entropy loss, while the optical flow loss is measured using the endpoint error (EPE).
The overall loss function is computed as a weighted sum of these three individual losses.

When the observed robot appears in the FOV of the observer robot, the mutual localization result ${T}_{ref,i}$ can be obtained as close to the ground truth as possible through the iterative rendering process mentioned above.
The planar movement of mobile robots is ideally represented in SE(2) for their poses.
We still estimate its SE(3) transformation because this enhances robustness against disturbances in the real-world environment and erroneous results can be eliminated by the pitch and roll angles.
However, due to the symmetry design of the mobile robot, there are still certain errors in estimating the yaw angle, which can be further improved in accuracy through global optimization with multiple frame observations.
\subsection{Pose Graph Optimization}
\label{PGO}
\begin{algorithm}[t]
	\caption{Pose Graph Optimization}
	\label{pose_graph_a}
	\KwIn{Refined robots mutual localization and odometry sets ${\mathcal{O}_{A}},{\mathcal{O}_{B}}$}
	\KwOut {Robots pose sets $\mathcal{A},\mathcal{B}$}
	{$\mathcal{N}_{A},\mathcal{N}_{B} \gets \texttt{DeleteErroneousNodes}(\mathcal{O}_{A},\mathcal{O}_{B})$};\\
	{$\mathcal{G}(\mathcal{V},\mathcal{E}) \gets \texttt{ConstructPoseGraph}(\mathcal{N}_{A},\mathcal{N}_{B})$};\\
	{${\Omega} \gets \texttt{CalculateInformationMatrix}(\mathcal{G})$};\\
	{$\mathcal{A},\mathcal{B} \gets \texttt{GlobalOptimization}(\Omega,\mathcal{G})$};\\
	\Return $\mathcal{A},\mathcal{B}$;
\end{algorithm}

When the observed robot becomes visible, it tends to remain within the FOV for a certain duration.
However, during the initial stages of robot rendezvous, it is common to encounter incomplete observations, which consequently leads to a decreased accuracy in network prediction.
Thus, leveraging multi-frame information for global optimization holds significant value.
Meanwhile, when the robot has relatively accurate odometry data, fusing it with the visual information mentioned above can generate higher accuracy mutual localization.
Therefore, PGO is employed to refine poses globally.

\begin{figure}[t]
	\centering
	\includegraphics[height=2in]{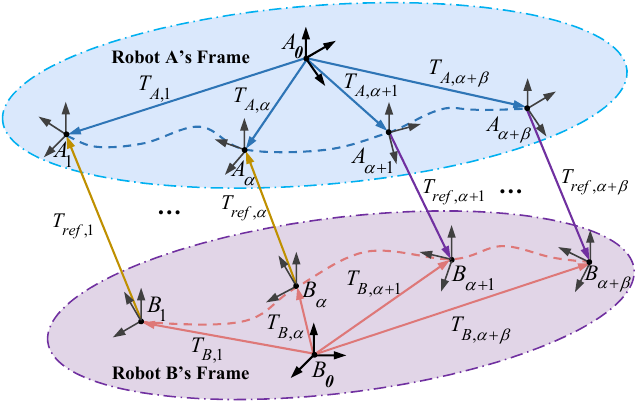}
	\caption{The coordinates of two robots during the rendezvous period. The yellow arrow indicates that Robot A was observed, totaling ${\alpha}$ times. Similarly, the purple arrow represents that Robot B was observed, amounting to ${\beta}$ times.}
	\label{pose_graph}
	\vspace{-4mm}
\end{figure}

Generally, the odometry of robot $A$ and robot $B$ at the $i$-th rendezvous are denoted as ${T}_{A,i}$ and ${T}_{B,i}$, respectively.
We define ${\mathcal{G}_{A,i}}=\{{{T}_{ref,i}},{{T}_{A,i}},{{T}_{B,i}}\}$, where ${T}_{ref,i}$ represents the coordinate transformation of the observed robot $A$ relative to the observer robot $B$, computed through the IR.
${\mathcal{G}_{B,i}}$ is defined in the similar way, except that ${T}_{ref,i}$ represents the pose of the observed robot $B$.
As illustrated in Fig.~\ref{pose_graph}, over a period of time, we assume that the initial poses of robot $A$ and $B$ are ${A}_{0}$ and ${B}_{0}$, respectively.
They have been observed ${\alpha}$ and ${\beta}$ times, and the corresponding observations are denoted by sets ${\mathcal{O}_{A}}=\{{\mathcal{G}_{A,1},\cdots,\mathcal{G}_{A,\alpha}}\}$ and ${\mathcal{O}_{B}}=\{{\mathcal{G}_{B,\alpha+1},\cdots,\mathcal{G}_{B,\alpha+\beta}}\}$.
Based on the mentioned definitions, Algorithm \ref{pose_graph_a} is employed to implement PGO, with some functions explained following.

\begin{itemize}[leftmargin=*]
	\item[$\bullet$]
	\texttt{DeleteErroneousNodes} is utilized to eliminate mutual localization results that exhibit significant errors, which commonly arise from incomplete observations.
	Due to the constraints of nearly planar motion of the mobile robot, the roll angle and pitch angle are both approximated to be 0.
	The rotation matrix ${R}_{ref,i}$ is extracted from ${T}_{ref,i}$ and converted into Euler angle, denoted as ${\hat{\vartheta }_{i}}$, ${\hat{\varphi }_{i}}$, and ${\hat{\psi }_{i}}$.
	If either ${\hat{\vartheta }_{i}}$ or ${\hat{\varphi }_{i}}$ exceeds the threshold $\sigma$, it indicates an error in the network, leading to the removal of the corresponding result.
	Finally, the selected optimized pose sets $\mathcal{N}_{A}=\{{\mathcal{G}_{A,1},\cdots,\mathcal{G}_{A,\alpha_d}}\}$ and $\mathcal{N}_{B}=\{{\mathcal{G}_{B,\alpha_d+1},\cdots,\mathcal{G}_{B,\alpha_d+\beta_d}}\}$ are obtained, where ${{\alpha }_{d}}\le \alpha $ and ${{\beta}_{d}}\le \beta $.
	\item[$\bullet$]
	\texttt{ConstructPoseGraph} integrates all the information of nodes and edges into the pose graph.
	The initial pose and poses calculated at the rendezvous points of each robot are employed as the node set $\mathcal{V}$, with the size of $2(1+{{\alpha }_{d}}+{{\beta}_{d}})$.
	The odometry and mutual localization results serve as the edge set $\mathcal{E}$, with the size of $3({{\alpha }_{d}}+{{\beta}_{d}})$.
	Thus, the pose graph $\mathcal{G}(\mathcal{V},\mathcal{E})$ is obtained.
	\item[$\bullet$]
	\texttt{CalculateInformationMatrix} quantifies the uncertainty of each degree of freedom (DoF) for the two nodes connected by the edge.
	Since the pose of the mobile robot belongs to SE(2), a $3 \times 3$ diagonal matrix is adopted as the information matrix ${{\Omega }}$.
	Especially, the edges $\mathcal{E}$ are divided into two categories: one type $\mathcal{E}_{m}$ denotes mutual localization constraints and the other type $\mathcal{E}_{o}$ represents robot odometry constraints.
	For ${e}_{o} \in \mathcal{E}_{o}$, the diagonal elements of ${{\Omega}_{o}}$ are set to ${\kappa}=100$.
	Meanwhile, for ${e}_{m} \in \mathcal{E}_{m}$, the information quantity ${{{\kappa}}_{m}}$ is defined as follows:
	\begin{equation}
		\label{eq9}
		{{\kappa}_{m}}=\frac{{{\kappa}}}{2}\exp (-|{{\hat\vartheta }_{m}}|-| {{\hat\varphi }_{m}}|)\
	\end{equation}
    where ${\hat\vartheta }_{m}$ and ${\hat\varphi }_{m}$ are the estimated roll angle and pitch angle, respectively.
	Since the rotation error of the refined localization result is greater compared to translation error, the information matrix ${{\Omega}_{m}}$is:
	\begin{equation}
		\label{eq10}
		{{\Omega}_{m}}=diag({{{\kappa}_{m}}},{{{\kappa}_{m}}},{\tau {{\kappa}_{m}}})
		\
	\end{equation}
    where $\tau$ represents the weight of the angle information.
	\item[$\bullet$]
	\texttt{GlobalOptimization} applies global optimization to all localization results.
	The error vector $\bm {r}$ for edge ${e}$ is defined as the translation and rotation error between the two nodes.
	The overall objective function is defined as follows:
	\begin{equation}
		\label{eq11}
		\underset{\mathcal A,\mathcal B}{\mathop{\min }}\,\frac{1}{2}({{\sum\limits_{{e}_{o} \in \mathcal{E}_{o}}{||\Omega _{o}^{\frac{1}{2}}{{\bm r}_{o}}||_{2}}}}+{{\sum\limits_{{e}_{m} \in \mathcal{E}_{m}}{||\Omega _{m}^{\frac{1}{2}}{{\bm r}_{m}}||_{2}}}})
		\
	\end{equation}
    where ${\mathcal A = \{A_0,\cdots,A_{\alpha_{d}+\beta_{d}}\}}$ represents the set of poses for robot $A$, and the same applies to ${\mathcal B}$.
\end{itemize}

Therefore, during the rendezvous period, RHAML optimizes mutual localization results sequentially with Algorithm \ref{ReML_a}.
Firstly, AMLNet is utilized for IML, yielding ${P}_{init,i}$.
Subsequently, IR is employed to further optimize the localization result, resulting in ${T}_{ref,i}$.
Finally, more accurate poses of robots are obtained through global pose graph optimization.

\begin{algorithm}[t]
	\caption{RHAML calculation}
	\label{ReML_a}
	\KwIn{Image ${I}_{A,i}, {I}_{B,i}$, odometry ${T}_{A,i}, {T}_{B,i}$, model $\mathcal{M}$}
	\KwOut{Robots pose sets $\mathcal{A},\mathcal{B}$}
	\While{\textit{Robots in Motion}}
	{
		\For{$r$ in \{A,B\}}
		{
			$\hat{o}_{i}, {P}_{init, i} \gets \texttt{IML}({I}_{r,i})$;\\
			\If{$\hat{o}_{i} == 1$}
			{
				${T}_{ref, i} \gets \texttt{IR}({I}_{r,i}, {P}_{init, i}, \mathcal{M})$;\\
				$\mathcal{O}_{r} \gets \texttt{UpdateState}({T}_{ref, i}, {T}_{A,i}, {T}_{B,i})$;\\
				$\mathcal{A},\mathcal{B} \gets \texttt{PGO}(\mathcal{O}_{A},\mathcal{O}_{B})$;\\
			}
		}
	}
	\Return $\mathcal{A},\mathcal{B}$
\end{algorithm}

\section{Simulations and Experiments}
\subsection{Experimental Setup}
\subsubsection{Datasets}
The simulation dataset is generated using OpenGL.
Initially, the robot is placed in random poses within 10 Gazebo simulation environments to collect background images.
Subsequently, random poses are assigned to the robot in OpenGL, which ensures that the robot is within the FOV, and the roll angle and pitch angle are all set to zero.
To mitigate the influence of different camera height, we randomly sample the height in $\mathcal N(0,{{0.05}^{2}})$.
Rendering is performed leveraging the CAD model and randomly selected background images, resulting in 70,000 RGB images.
Meanwhile, 10,000 background images are added to the dataset as the robot does not exist, with all DoF set to 0.
80,000 images along with their corresponding poses are employed to train AMLNet.
To train the DeepIM, based on the approximate prediction error range of AMLNet, Gaussian noises are added to the ground truth as initial pose inputs across three DoF: $\Delta x\sim \mathcal N(0,{{0.15}^{2}})$, $\Delta y\sim \mathcal N(0,{{0.15}^{2}})$, $\Delta \psi \sim \mathcal N(0,{{20}^{2}})$, where the standard deviations are 0.15 m, 0.15 m and 20 degrees, respectively.
\subsubsection{Implementation Details}
The Pioneer3-DX robot and the Turtlebot3 Burger robot are used for simulation and experiment, respectively.
Due to the significant differences in their characteristics, we train separate CNN models for each.
The hardware platform consisting of an Intel i7 3.7GHz CPU and a GTX 3060 Ti GPU.
We implement the AMLNet using PyTorch, with input image normalization to a size of $224\times224$.
For AIncep blocks, we adopt three unidirectional convolutional kernels of each dimension in the ACN, with sizes of 3, 5, and 7.
${\omega }_{\psi }$ and ${\omega }_{p}$ are set to 3 and 2, respectively.
During training, we set the batch size to 32.
AdamW optimizer is employed to minimize the loss function ${L}_{A}$, with an initial learning rate of 0.001, which is adjusted using the cosine annealing strategy.
For training the DeepIM, the number of iterations is set to 4, while the remaining parameters are the same as in \cite{li2018deepim}.
For PGO, we set the angle threshold $\sigma = 10$ degrees and $\tau = 0.5$.
\subsection{Mutual Localization Performance}
\begin{figure}[t]
	\centering
	\includegraphics[height=1.8in]{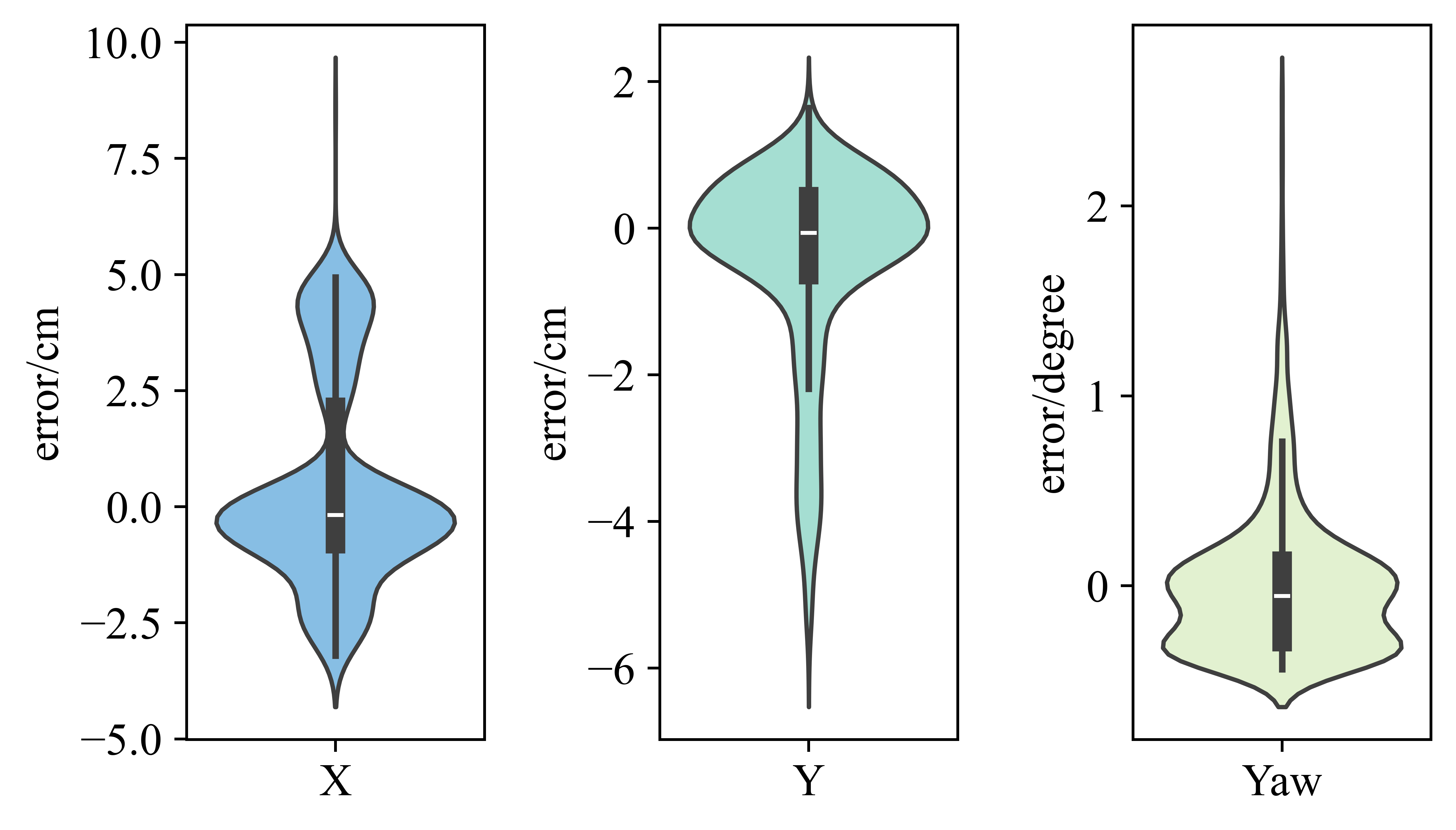}
	\vspace{-3mm}
	\caption{The violin plot of pose errors at 1000 points with RHAML.}
	\label{violin}
	\vspace{-4mm}
\end{figure}
RHAML utilizes images captured by the observer robot at 1000 rendezvous points, as shown in Fig.~\ref{comparision}, for mutual localization and calculates the relative pose error at each point.
The violin plot displaying the error distributions for each DoF is depicted in Fig.~\ref{violin}.
It can be observed that the mean errors for all DoF are close to zero, and the variances are small to achieve high-precision mutual localization.
Meanwhile, the error distribution between each DoF is relatively uniform without significant difference, indicating that RHAML has achieved excellent performance in balancing localization accuracy across all DoFs.
There are a few outliers present in all DoFs, which can be attributed to blurry image rendering when the observed robot is situated at a considerable distance.
\subsection{Comparison With Other Methods}
\begin{figure}[!t]
	\centering  
	\subfigure[]{ 
		\centering    
		\includegraphics[height=1in]{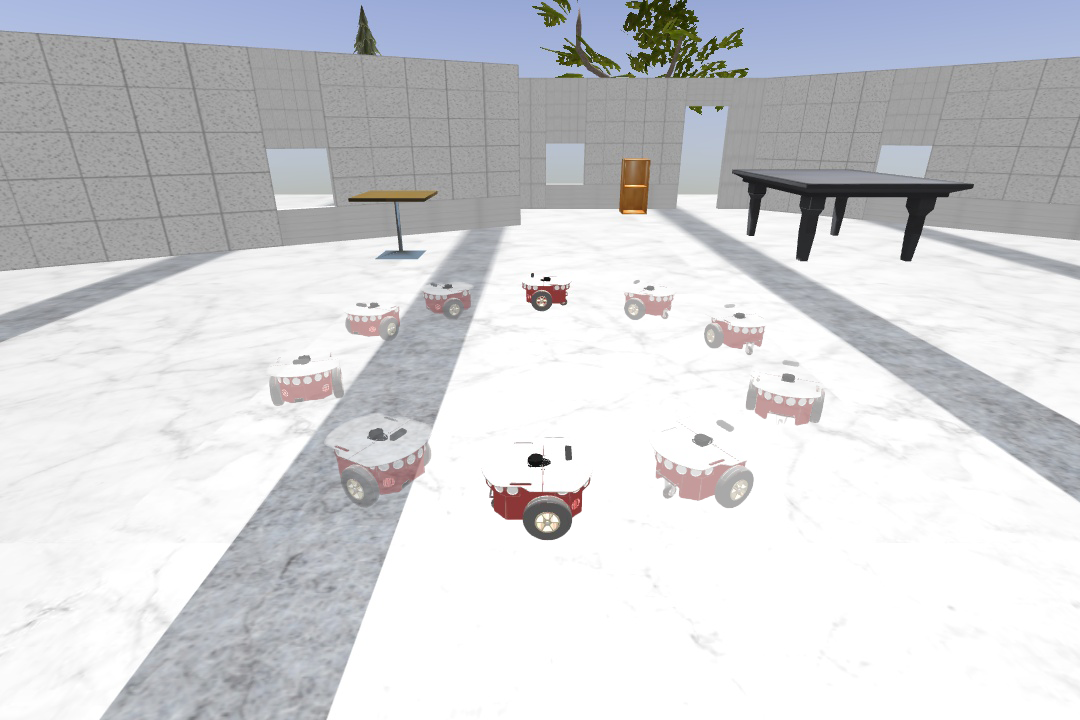}
		\label{circle1}
	}
	\subfigure[]{  
		\centering    
		\includegraphics[height=1in]{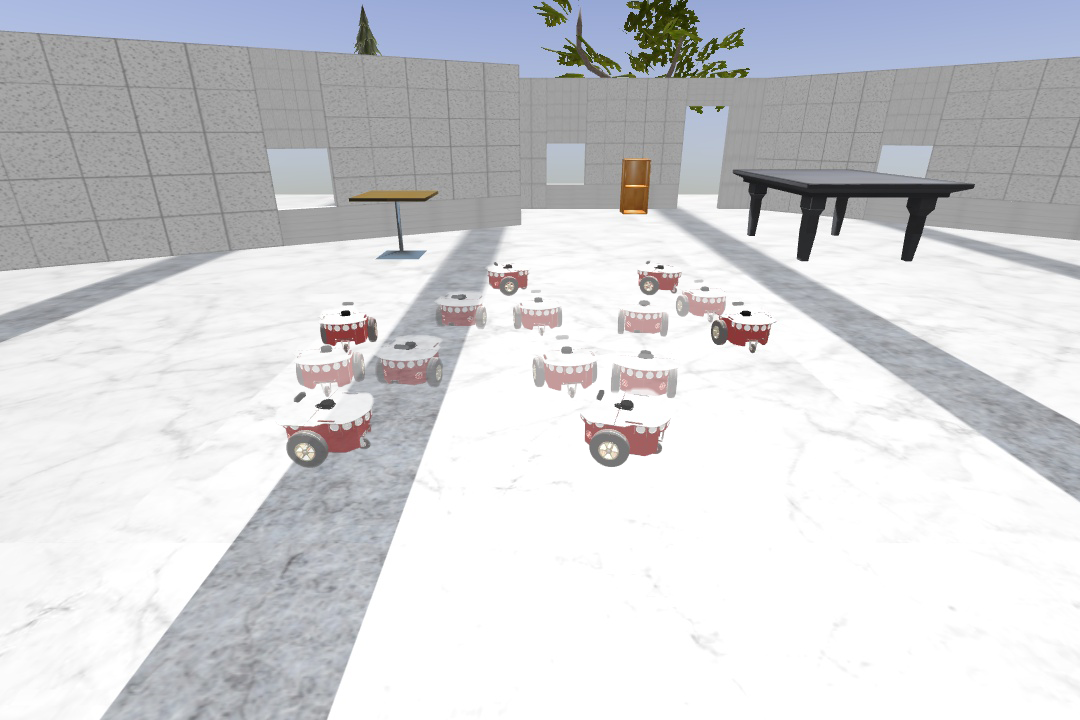}
		\label{sin1}
	}
	\subfigure[]{   
		\centering    
		\includegraphics[height=1.15in]{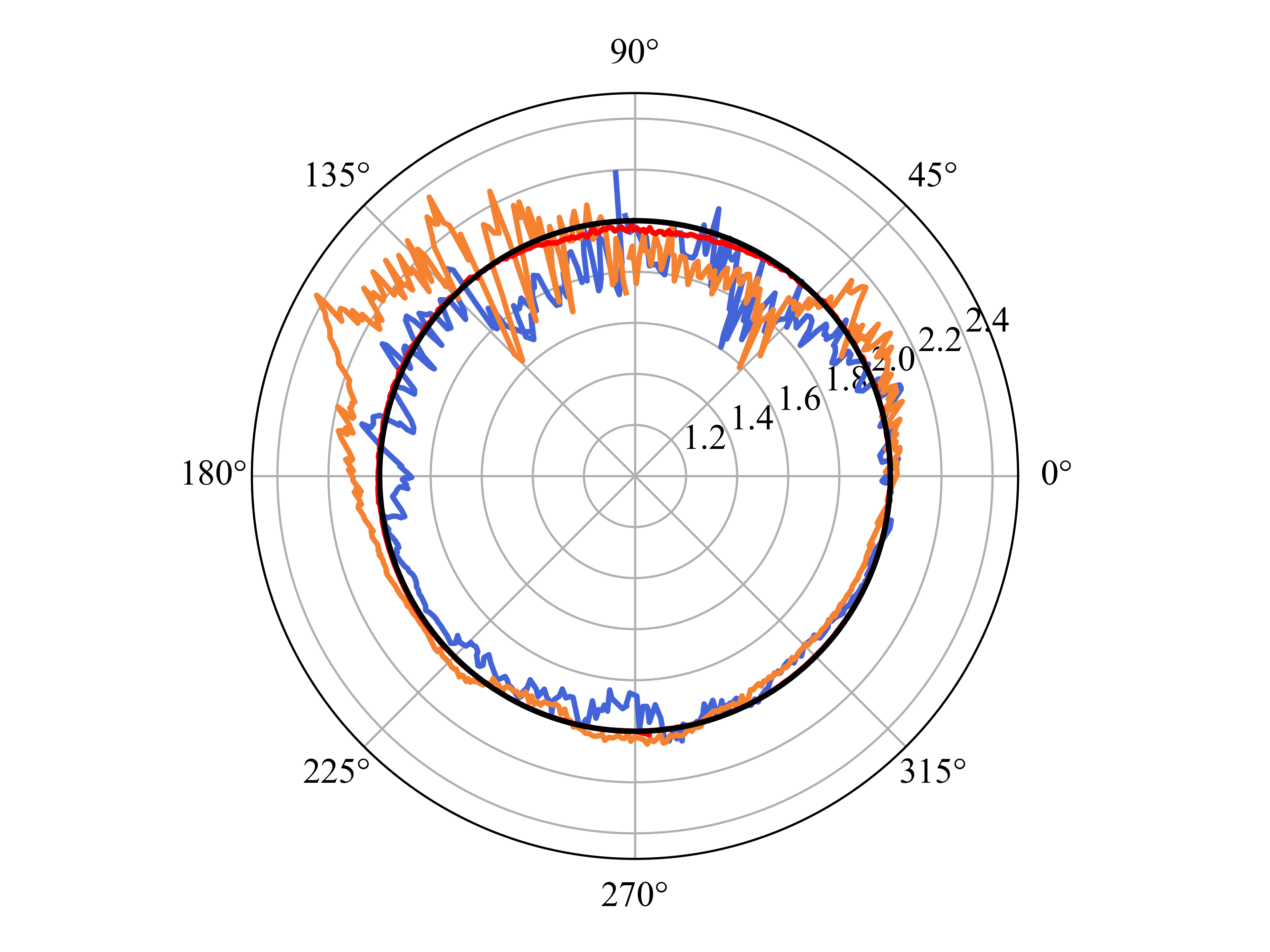}
		\label{circle2}
	}
	\subfigure[]{   
		\centering    
		\includegraphics[height=1.15in]{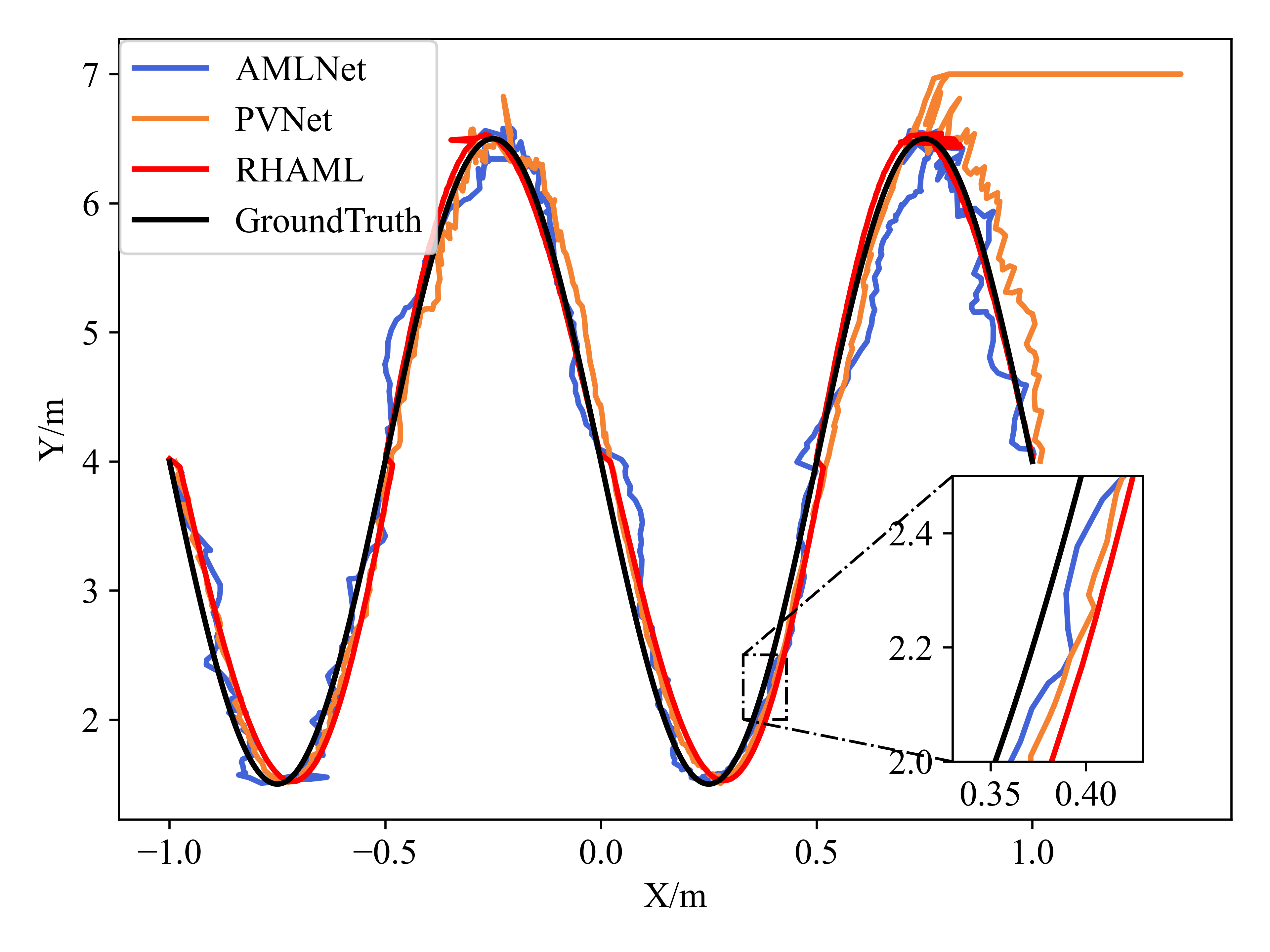}
		\label{sin2}
	}
	\caption{The observed robot moves on two curves and the observer robot estimates pose at each rendezvous point with different methods. (a), (b) The observed robot moves on a circle and a sinusoid respectively. (c), (d) Comparison of three methods for mutual localization on these two curves.}    
	\label{comparision}
	\vspace{-4mm}
\end{figure}

To validate the algorithm's ability for mutual localization between robots during continuous motion with a wide-depth-range, we conduct two motion scenes.
Firstly, the observed robot is instructed to navigate in a circular path around the center located 4.5 m directly in front of the observer robot, with a radius of 2 m, as shown in Fig.~\ref{circle1}.
Then, the observed robot follows a sinusoidal trajectory, as depicted in Fig.~\ref{sin1}, with an amplitude of 2.5m.
In total, we collect 1000 images to facilitate the mutual localization between the robots.

As baseline methods, FrontNet\cite{DL_Rendezvous2}, DOPE\cite{DOPE}, and PVNet \cite{PVNet} are all retrained with our dataset.
Visual transformers typically segment the image into patches to capture contextual information, but they struggle with handling local features.
While feature pyramid networks leverage information from multiple feature maps, their scale variation is limited.
These two backbone networks are not compared with ours.
The simulation results are presented in Table \ref{table_comparision}, where the errors for each DoF are represented by the mean and variance.
Meanwhile, the absolute trajectory error (ATE) is calculated with Eq. \eqref{eq12} to demonstrate the overall performance of mutual localization.
The model's parameters (Params) and Multiply-Accumulate Operations (MACs) are also compared.
\begin{equation}
	\label{eq12}
	{\rm ATE}=\sqrt{\frac{1}{N_r}\sum\limits_{i=1}^{N_r}{\left\| \log {{(T_{gt,i}^{-1}{{T}_{est,i}})}^{\vee }} \right\|_{2}^{2}}}\
\end{equation}
where ${N_r}$ represents the number of rendezvous points, i.e. 1000, and $T_{gt,i}$ and ${T}_{est,i}$ denote the ground truth and estimated result of mutual localization, respectively.
\begin{table}[t]
	\centering
	\fontsize{7.5}{11}\selectfont
	\caption{Comparison of Translation and Rotation Error (Mean $\pm $ Std) at 1000 Rendezvous Points with Different Methods}
	\vspace{-2mm}
	\label{table_comparision}
	\begin{threeparttable}
		\begin{tabular}{@{}c@{\hspace{5pt}}c@{\hspace{5pt}}c@{\hspace{5pt}}c@{}c@{}c@{}c@{}}
			\toprule
			\multirow{2}{*}{Methods}&
			\multicolumn{1}{c}{$\left|{\Delta }_{x} \right|$} &\multicolumn{1}{c}{$\left| {\Delta }_{y} \right|$}&\multicolumn{1}{c}{$\left|{\Delta }_{\psi }\right|$}&\multicolumn{1}{c}{${\rm{ATE}}$}&\multicolumn{1}{c}{${\rm{Params}}$}&\multicolumn{1}{c}{${\rm{MACs}}$}\\
			&(cm)&(cm)&(degree)&&(M)&(G)\\
			\midrule
			FrontNet &30.3$\pm $36.7 & 50.9$\pm $75.8  & 64.9$\pm $53.2 & 2.21 & 0.33 & 0.07 \\
			DOPE & 20.4$\pm $27.8 & 28.8$\pm $23.3  & 25.4$\pm32.3 $ & 1.02 & 50.3 & 205 \\
			PVNet &  4.4$\pm $5.4  &10.8$\pm $18.6 & 5.9$\pm22.1 $ & 0.37& 13.0 & 72.8  \\
			\midrule
			AMLNet & 2.9$\pm $2.3 & 7.3$\pm $6.6& 6.8$\pm12.9 $ & 0.29& 49.8 & 8.8  \\
			RHAML &  {\textbf {1.6$\pm $1.5}}&  {\textbf {0.9$\pm $1.1}}  &  {\textbf{0.4$\pm $0.4}} & {\textbf{0.03}} & 109 & 50.7 \\
			\bottomrule
		\end{tabular}
	\end{threeparttable}
	\vspace{-2mm} 
\end{table}

Because of its lightweight architecture and minimal MACs, FrontNet achieves fastest inference speed, but its feature extraction capability is limited, leading to relatively poor mutual localization performance.
DOPE enables mutual localization but with limited accuracy, possibly due to inaccurate corner prediction of the robot.
PVNet performs exceptionally well in predicting yaw angles and also shows significant improvement in translation prediction.
Compared with our AMLNet, PVNet exhibits a smaller average error in yaw angle but a larger in translation.
Moreover, as shown in Fig.~\ref{sin2}, when the robot is far away, PVNet shows some outliers, which we artificially limit to 7 m.
This might be attributed to fewer pixels corresponding to the robot at longer distances, leading to inaccurate keypoint voting results.
Meanwhile, PVNet has a larger ATE due to the significant variance in yaw angle caused by outliers, affecting overall accuracy.
However, it is undeniable that PVNet performs better in estimating the poses of nearby robots.
Finally, our RHAML achieves the highest precision in all three DoFs.
Specifically, compared to PVNet, which performs the best among baseline methods, RHAML improves the accuracy of predicting $x$, $y$, and $\psi$ by 2.8 cm, 9.9 cm, and 5.5 degree, respectively.
As a result, the ATE is improved by 0.34, which highlights the advantages of anisotropic convolution, as well as local and global refinement.
The predictions of three methods at each rendezvous point are depicted in Fig.~\ref{circle2} and Fig.~\ref{sin2}, wherein it is noticeable that our method closely aligns with the ground truth.
\subsection{Ablation Study}
\begin{table}[t]
	\centering
	\fontsize{8}{11}\selectfont
	\caption{Ablation Study for Anisotropic Convolution in AMLNet}
	\vspace{-2mm}
	\label{ablation_ACN}
	\begin{threeparttable}
		\begin{tabular}{lcccc}
			\toprule
			\multirow{1}{*}{ACN}&
			\multicolumn{1}{c}{$\left|{\Delta }_{x} \right|/\rm cm$} &\multicolumn{1}{c}{$\left| {\Delta }_{y} \right|/\rm cm$}&\multicolumn{1}{c}{$\left|{\Delta }_{\psi}\right|/\rm degree$}&\multicolumn{1}{c}{${\rm{ATE}}$}\cr
			\midrule
			{$\mathcal{K}= \varnothing $} &{8.3$\pm $8.2} & {13.9$\pm $12.6} & {14.7$\pm $18.0}&0.70 \\
			$\mathcal{K}=\{3\}$ & 15.7$\pm $8.9 & 14.4$\pm $12.8 & 11.0$\pm$10.8&0.64  \\
			$\mathcal{K}=\{3, 5\}$ & 5.2$\pm $5.8 & 9.6$\pm $9.0 & 11.3$\pm8.6 $ &0.45\\
			$\mathcal{K}=\{3, 5, 7\}$ & {\textbf{2.9$\pm $2.3}} & {\textbf{7.3$\pm $6.6}}& {\textbf{6.8$\pm$12.9}} & $\textbf{0.29}$ \\
			\bottomrule
		\end{tabular}
	\end{threeparttable}
\end{table}
\begin{table}
	\centering
	\fontsize{8}{11}\selectfont
	\caption{Ablation Study for RHAML Architecture}
	\vspace{-2mm}
	\label{ablation_architecture}
	\begin{tabular}{@{\hspace{4pt}}c@{\hspace{4pt}}c@{\hspace{4pt}}c@{\hspace{4pt}}|ccc@{\hspace{5pt}}c@{\hspace{5pt}}c}
		\Xhline{1pt}
		IML & IR & PGO & {$\left|{\Delta }_{x} \right|/\rm cm$} &{$\left|{\Delta }_{y} \right|/\rm cm$} & {$\left|{\Delta }_{\psi} \right|/\rm degree$}&{${\rm{ATE}}$}&{${\rm{FPS}}$}\\
		\Xcline{1-1}{0.4pt}
		\Xhline{0.5pt}
		{\ding{52}} & {\ding{56}} & {\ding{56}} & 2.9$\pm $2.3 & 7.3$\pm $6.6& 6.8$\pm12.9 $ & 0.29 & $\textbf{34.5}$ \\
		{\ding{52}} & {\ding{52}} & {\ding{56}} & {\textbf{1.3$\pm $1.1}} & {4.9$\pm $7.7} & {2.9$\pm $14.8}& 0.30& 7.7 \\
		{\ding{52}} & {\ding{56}} & {\ding{52}} & {6.1$\pm $4.1} & {13.0$\pm $4.1} & {2.6$\pm $3.8}& 0.17 & 2.2 \\
		{\ding{52}} & {\ding{52}} & {\ding{52}} &  { {1.6$\pm $1.5}}&  {\textbf {0.9$\pm $1.1}}  &  {\textbf{0.4$\pm $0.4}} & {\textbf{0.03}}& 1.8 \\		
		\Xhline{1pt}
	\end{tabular}
	\vspace{-2mm} 
\end{table}
\subsubsection{Effect of Anisotropic Convolution}
An ablation experiment is performed to validate the adaptability of ACN to the large scale variations, and the AMLNet results are presented in Table \ref{ablation_ACN}.
We adopt four different ACN modules in total, with $\mathcal{K}= \varnothing$ representing the use of a regular $3\times3$ convolutional kernel.
Through comparison, it can be observed that when using $\mathcal{K}=\{3, 5, 7\}$, AMLNet exhibits the highest precision in three DoFs.
This can be explained as when the observed robot is far away, the learned weight of the small convolution kernel is significant to extract finer local features.
On the contrary, when the robot is close, the weight of the large convolution kernel becomes prominent, ensuring that the robot's entire features are essentially covered after multiple convolutions.
Compared to the fixed receptive field, it reduces the prediction errors in $x$, $y$, and $\psi$ by 5.4 cm, 6.6 cm, and 7.9 degrees, respectively.
The ATE decreases by 0.41.
Thus, this indicates that the ACN module is capable of providing the flexible receptive field to accommodate the large scale variations.
\subsubsection{Architectural Components}
Another ablation experiment is conducted to validate the necessity of the proposed modules within the overall algorithm architecture.
Meanwhile, the time efficiency analysis of RHAML in various usage modes is also included.
The results of this experiment are recorded in Table \ref{ablation_architecture}.
IML can achieve fast initial localization at a speed of 34.5 frames per second (FPS).
Compared to using only IML, the combination of IML and IR results in a notable improvement in accuracy, particularly in the prediction of translation, with the most accurate predictions observed in the $x$ direction.
However, due to the increased yaw angle variance, ATE actually increases by 0.01.
This is because ATE is more sensitive to angle errors.
Additionally, the simultaneous usage of IML and PGO significantly reduces rotation error.
Although the translation error increase after incorporating global refinement, the mean and variance of the angle error significantly decrease, resulting in a reduction in ATE.
Ultimately, when all three modules are simultaneously employed, an improvement in prediction accuracy is observed for $x$, $y$, and $\psi$ compared to the usage of only IML, with enhancements by a factor of 1.8, 8.1, and 17.0, achieving the best performance in overall prediction.
Due to the impact of PGO, RHAML achieves a speed of 1.8 FPS.
However, PGO is executed only once for multiple observations, allowing its time to be averaged per frame.
Since the number of observations is variable, in Table \ref{ablation_architecture}, we accumulate the time of PGO for each frame.
Consequently, IML provides an initial estimation, IR optimizes translation prediction, and PGO greatly reduces rotation errors.
\subsection{Application}
\begin{figure}[t]
	\centering
	\subfigure[]{ 
		\centering   
		\includegraphics[height=1.5in]{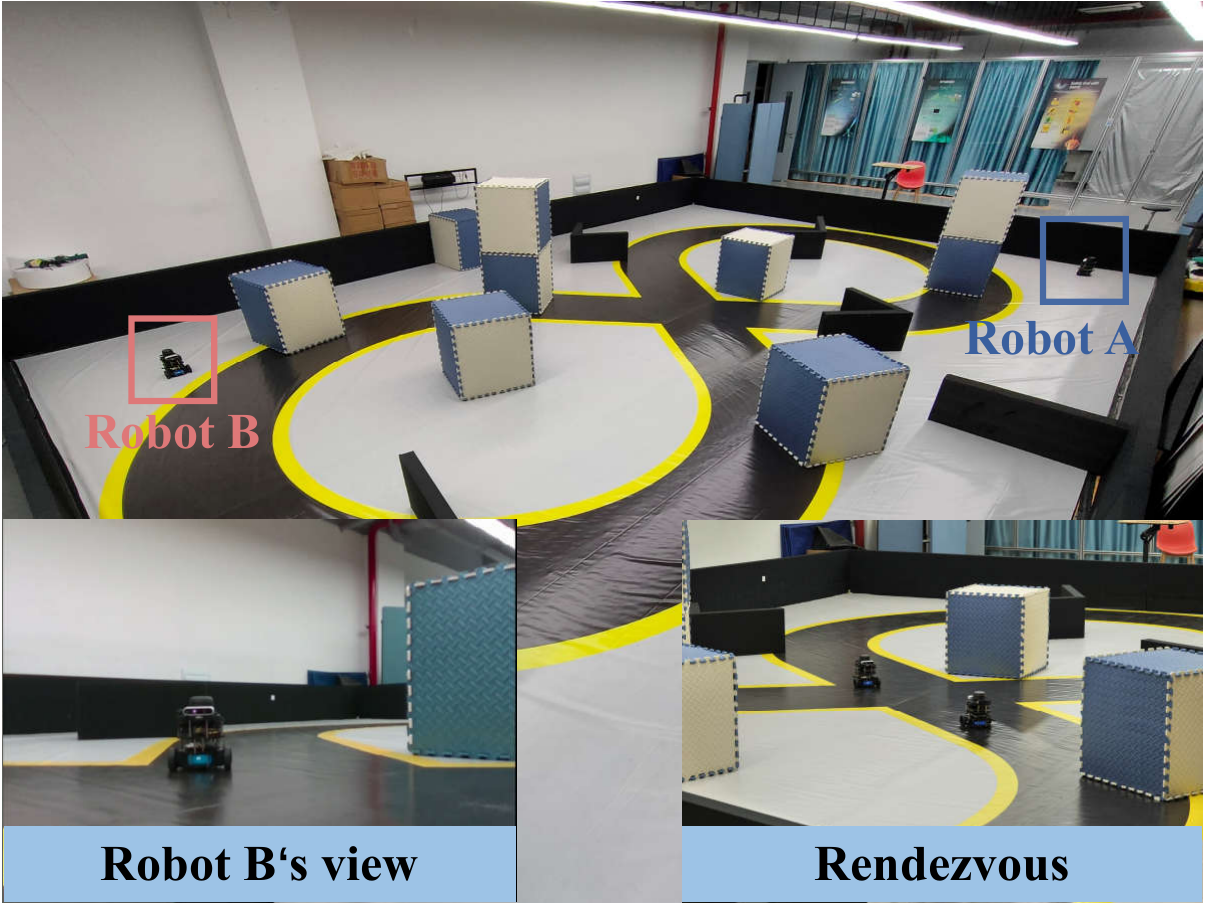}
		\label{fig_realworld_image}
	}
	\subfigure[]{   
		\centering    
		\includegraphics[height=1.5in]{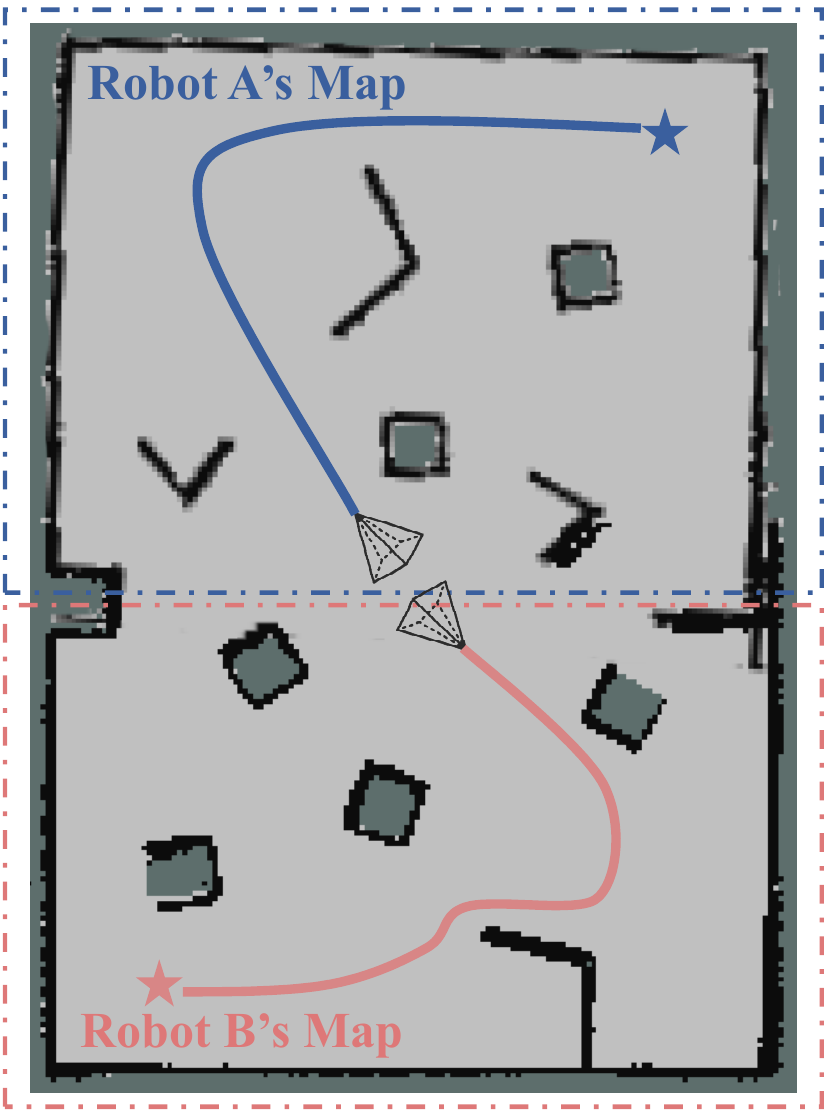}
		\label{fig_realworld_result}
	}
	\vspace{-2mm}
	\caption{Real-world experiment on multi-robot map fusion. (a) Experiment scenario and a snapshot of the rendezvous process. (b) Map fusion result.}   
	\label{fig_realworld}
	\vspace{-4mm}
\end{figure}
RHAML is applied to the collaborative exploration by two robots in an unknown environment and map fusion when they rendezvous, using Cartographer\cite{carto} for SLAM.
Assuming ${}^{{{A}_{0}}}{{T}_{{{B}_{0}}}}$ represents the pose of robot B's map origin ${B}_{0}$ in ${A}_{0}$, we compute it using the following fundamental relationship:
\begin{equation}
	\label{eq13}
	{}^{{{A}_{0}}}{{T}_{{{B}_{0}}}}={}^{{{A}_{0}}}{{T}_{{{A}_{i}}}}{}^{{{A}_{i}}}{{T}_{{{B}_{i}}}}{}^{{{B}_{i}}}{{T}_{{{B}_{0}}}}
\end{equation}
where ${}^{{{A}_{0}}}{{T}_{{{A}_{i}}}}$ and ${}^{{{B}_{i}}}{{T}_{{{B}_{0}}}}$ are obtained from the odometry, and ${}^{{{A}_{i}}}{{T}_{{{B}_{i}}}}$ is estimated through mutual localization with RHAML.

The experiment is conducted on a rectangular field with a size of $10\rm m\times7\rm m$, where obstacles are randomly placed as shown in Fig.~\ref{fig_realworld_image}.
Each robot is equipped with an RGB camera and a 2D LIDAR.
Robots share odometry data between each other.
Two robots initially explore unknown areas based on their frontiers and perform SLAM to construct their respective grid maps.
When they simultaneously reach the middle region, a rendezvous occurs.
The robot rendezvous lasts for about 5 seconds.
10 rendezvous points at equal time intervals are extracted and RHAML is employed to achieve mutual localization.
The maps generated by two robots are merged into a common reference frame with Eq. \eqref{eq13}.
The merged map and the trajectories of each robot are depicted in Fig.~\ref{fig_realworld_result}.
Due to the influence of sensor noise, slight deviation occurs during the map fusion.
The predicted transformation between the initial frames of two robots has an approximate rotation error of 0.9 degree and a translation error of 5 cm.
The increase in translation error is primarily due to the entanglement between rotation and translation in ${}^{{{A}_{0}}}{{T}_{{{A}_{i}}}}{}^{{{A}_{i}}}{{T}_{{{B}_{i}}}}{}^{{{B}_{i}}}{{T}_{{{B}_{0}}}}$, as well as the influence of different robot appearances.
However, the map generally aligns well with the physical layout of the scene, verifying the practical application ability of RHAML.
\subsection{Discussion}
In general, RHAML utilizes visual observations and odometry to achieve high-precision mutual localization through a hierarchical framework.
Therefore, depending on the task requirements, RHAML provides different solutions.
When high accuracy is not required, the IML can be used alone, which is faster.
When precise translation is the only requirement, such as in multi-robot obstacle avoidance, the IML can be combined with the IR.
This approach allows achieving the desired result only through a single-frame visual observation, without the need for odometry.
For tasks that demand high-precision relative poses, such as multi-robot map fusion, the complete framework of RHAML can be employed.
Thus, RHAML offers a flexible architecture for mutual localization.

However, when the number of robots within the FOV exceeds one, RHAML only performs mutual localization with one of the robots.
Therefore, when there are more than two robots in the scene, incomplete localization may occur.
An effective approach is to incorporate anchor boxes in the IML, enabling multi-robot detection and prediction.
RHAML also relies on relatively accurate odometry data, which may pose challenges when used in more challenging environments.
\section{Conclusion}
This paper proposes a hierarchical marker-less multi-robot mutual localization framework, called RHAML, taking RGB images and odometry as inputs.
The initial predictions are obtained with AMLNet, followed by the IR and PGO to sequentially refine the localization results.
Anisotropic convolution provides the network with flexible receptive fields, which can offer better initial values for subsequent optimization.
Benefiting from RHAML's hierarchical architecture, it provides a flexible framework selection for different requirements.
Even with the large scale variations, RHAML achieves an average translation error of less than 2 cm and rotation error of less than 0.5 degrees in multi-robot mutual localization.

Regarding future work, on the one hand, enhancing AMLNet's capability to simultaneously recognize and localize multi-robots within the FOV of the observer robot will make the entire system more robust.
On the other hand, it is also promising to explore methods for expediting the rendezvous of multi-robots in an unknown environment.
\bibliographystyle{ieeetr}
\bibliography{ref}

\begin{thebibliography}{10}

\bibitem{CMU_explor}
C.~Cao, H.~Zhu, Z.~Ren, H.~Choset, and J.~Zhang, ``Representation granularity
  enables time-efficient autonomous exploration in large, complex worlds,''
  {\em Sci. Robot.}, vol.~8, no.~80, p.~eadf0970, 2023.

\bibitem{Pasqualetti}
F.~Pasqualetti, A.~Franchi, and F.~Bullo, ``{On Cooperative Patrolling:
  Optimal} trajectories, complexity analysis, and approximation algorithms,''
  {\em {IEEE} Trans. Robotics}, vol.~28, no.~3, pp.~592--606, 2012.

\bibitem{Hu}
J.~Hu, W.~Liu, H.~Zhang, J.~Yi, and Z.~Xiong, ``Multi-robot object transport
  motion planning with a deformable sheet,'' {\em {IEEE} Robot. Automat.
  Lett.}, vol.~7, no.~4, pp.~9350--9357, 2022.

\bibitem{Dong}
S.~Dong, K.~Xu, Q.~Zhou, A.~Tagliasacchi, S.~Xin, M.~Nie\ss{}ner, and B.~Chen,
  ``Multi-robot collaborative dense scene reconstruction,'' {\em ACM Trans.
  Graph.}, vol.~38, jul 2019.

\bibitem{Gaofei2022}
Y.~Wang, X.~Wen, L.~Yin, C.~Xu, Y.~Cao, and F.~Gao, ``Certifiably optimal
  mutual localization with anonymous bearing measurements,'' {\em {IEEE} Robot.
  Automat. Lett.}, vol.~7, no.~4, pp.~9374--9381, 2022.

\bibitem{Zhou}
X.~S. Zhou and S.~I. Roumeliotis, ``Determining 3-d relative transformations
  for any combination of range and bearing measurements,'' {\em {IEEE} Trans.
  Robotics}, vol.~29, no.~2, pp.~458--474, 2013.

\bibitem{Liu}
J.~Liu and G.~Hu, ``Relative localization estimation for multiple robots via
  the rotating {Ultra}-{Wideband} tag,'' {\em {IEEE} Robot. Automat. Lett.},
  vol.~8, no.~7, pp.~4187--4194, 2023.

\bibitem{Arcuo_1}
Y.~Jang, C.~Oh, Y.~Lee, and H.~J. Kim, ``Multirobot collaborative monocular
  slam utilizing rendezvous,'' {\em {IEEE} Trans. Robotics}, vol.~37, no.~5,
  pp.~1469--1486, 2021.

\bibitem{CCM-SLAM}
P.~Schmuck and M.~Chli, ``{CCM-SLAM}: Robust and efficient centralized
  collaborative monocular simultaneous localization and mapping for robotic
  teams,'' {\em J. Field Robotics}, vol.~36, no.~4, pp.~763--781, 2019.

\bibitem{TopoMap}
Z.~Zhang, J.~Yu, J.~Tang, Y.~Xu, and Y.~Wang, ``{MR-TopoMap}: Multi-robot
  exploration based on topological map in communication restricted
  environment,'' {\em {IEEE} Robot. Automat. Lett.}, vol.~7, no.~4,
  pp.~10794--10801, 2022.

\bibitem{Kimera-Multi}
Y.~Tian, Y.~Chang, F.~Herrera~Arias, C.~Nieto-Granda, J.~P. How, and
  L.~Carlone, ``{Kimera-Multi}: Robust, distributed, dense metric-semantic slam
  for multi-robot systems,'' {\em {IEEE} Trans. Robotics}, vol.~38, no.~4,
  pp.~2022--2038, 2022.

\bibitem{CNN_trans}
J.~Kim, D.-S. Han, and B.-T. Zhang, ``Robust map fusion with visual attention
  utilizing multi-agent rendezvous,'' in {\em Proc. IEEE Int. Conf. Robot.
  Autom.}, pp.~2062--2068, 2023.

\bibitem{DL_Rendezvous2}
S.~Bonato, S.~C. Lambertenghi, E.~Cereda, A.~Giusti, and D.~Palossi,
  ``Ultra-low power deep learning-based monocular relative localization onboard
  nano-quadrotors,'' in {\em Proc. IEEE Int. Conf. Robot. Autom.},
  pp.~3411--3417, 2023.

\bibitem{DOPE}
J.~Tremblay, T.~To, B.~Sundaralingam, Y.~Xiang, D.~Fox, and S.~Birchfield,
  ``Deep object pose estimation for semantic robotic grasping of household
  objects,'' in {\em Proc. Conf. Robot. Learn.}, 2018.

\bibitem{Bultmann}
S.~Bultmann, R.~Memmesheimer, and S.~Behnke, ``External camera-based mobile
  robot pose estimation for collaborative perception with smart edge sensors,''
  in {\em Proc. IEEE Int. Conf. Robot. Autom.}, pp.~8194--8200, 2023.

\bibitem{li2018deepim}
Y.~Li, G.~Wang, X.~Ji, Y.~Xiang, and D.~Fox, ``{DeepIM: Deep} iterative
  matching for 6d pose estimation,'' {\em Int. J. Comp. Vis.}, vol.~128, no.~3,
  pp.~657--678, 2020.

\bibitem{Semanticgeometric}
S.~Garg, N.~Suenderhauf, and M.~Milford, ``Semantic–geometric visual place
  recognition: a new perspective for reconciling opposing views,'' {\em Int. J.
  Robot. Res.}, vol.~41, no.~6, pp.~573--598, 2022.

\bibitem{AutoMerge}
P.~Yin, S.~Zhao, H.~Lai, R.~Ge, J.~Zhang, H.~Choset, and S.~Scherer,
  ``{AutoMerge}: A framework for map assembling and smoothing in city-scale
  environments,'' {\em {IEEE} Trans. Robotics}, vol.~39, no.~5, pp.~3686--3704,
  2023.

\bibitem{Arcuo_2}
P.~Zhang, G.~Chen, Y.~Li, and W.~Dong, ``Agile formation control of drone
  flocking enhanced with active vision-based relative localization,'' {\em
  {IEEE} Robot. Automat. Lett.}, vol.~7, no.~3, pp.~6359--6366, 2022.

\bibitem{Gaofei2023}
Y.~Wang, X.~Wen, Y.~Cao, C.~Xu, and F.~Gao, ``Bearing-based relative
  localization for robotic swarm with partially mutual observations,'' {\em
  {IEEE} Robot. Automat. Lett.}, vol.~8, no.~4, pp.~2142--2149, 2023.

\bibitem{CNN_xyzBouning1}
F.~Schilling, F.~Schiano, and D.~Floreano, ``Vision-based drone flocking in
  outdoor environments,'' {\em {IEEE} Robot. Automat. Lett.}, vol.~6, no.~2,
  pp.~2954--2961, 2021.

\bibitem{CNN_xyzBouning2}
M.~Vrba and M.~Saska, ``Marker-less micro aerial vehicle detection and
  localization using convolutional neural networks,'' {\em {IEEE} Robot.
  Automat. Lett.}, vol.~5, no.~2, pp.~2459--2466, 2020.

\bibitem{DL_Rendezvous1}
S.~Li, C.~De~Wagter, and G.~C. H.~E. De~Croon, ``Self-supervised monocular
  multi-robot relative localization with efficient deep neural networks,'' in
  {\em Proc. IEEE Int. Conf. Robot. Autom.}, pp.~9689--9695, 2022.

\bibitem{UAV_CNN}
L.~Crupi, A.~Giusti, and D.~Palossi, ``High-throughput visual nano-drone to
  nano-drone relative localization using onboard fully convolutional
  networks,'' {\em arXiv preprint arXiv:2402.13756}, 2024.

\bibitem{PVNet}
S.~Peng, X.~Zhou, Y.~Liu, H.~Lin, Q.~Huang, and H.~Bao, ``{PVNet}: Pixel-wise
  voting network for 6dof object pose estimation,'' {\em {IEEE} Trans. Pattern
  Anal. Machine Intell.}, vol.~44, no.~6, pp.~3212--3223, 2022.

\bibitem{OnePose}
J.~Sun, Z.~Wang, S.~Zhang, X.~He, H.~Zhao, G.~Zhang, and X.~Zhou, ``{{OnePose}:
  One-Shot Object Pose Estimation without CAD Models},'' in {\em Proc. IEEE
  Conf. Comp. Vis. Pattern Recog.}, pp.~6815--6824, 2022.

\bibitem{CosyPose}
Y.~Labb{\'e}, J.~Carpentier, M.~Aubry, and J.~Sivic, ``{{CosyPose}: Consistent
  Multi-view Multi-object 6D Pose Estimation},'' in {\em Proc. Europ. Conf.
  Comp. Vis.}, pp.~574--591, Springer International Publishing, 2020.

\bibitem{Anisotropic}
J.~Li, P.~Wang, K.~Han, and Y.~Liu, ``Anisotropic convolutional neural networks
  for {RGB-D} based semantic scene completion,'' {\em {IEEE} Trans. Pattern
  Anal. Machine Intell.}, vol.~44, no.~11, pp.~8125--8138, 2022.

\bibitem{yu2023inceptionnext}
W.~Yu, P.~Zhou, S.~Yan, and X.~Wang, ``{InceptionNeXt: when inception} meets
  {convnext},'' {\em arXiv preprint arXiv:2303.16900}, 2023.

\bibitem{carto}
W.~Hess, D.~Kohler, H.~Rapp, and D.~Andor, ``Real-time loop closure in 2d lidar
  slam,'' in {\em Proc. IEEE Int. Conf. Robot. Autom.}, pp.~1271--1278, 2016.

\end{thebibliography}
\end{document}